\documentclass[conference]{IEEEtran}
\IEEEoverridecommandlockouts
\usepackage{cite}
\usepackage[numbers]{natbib}

\usepackage{amsmath,amssymb,amsfonts}
\usepackage{algorithm}
\usepackage{algpseudocode}
\usepackage{graphicx}
\usepackage{textcomp}
\usepackage{xcolor}
\usepackage{adjustbox}
\usepackage{booktabs}
\usepackage{multicol, multirow}
\usepackage{makecell}
\usepackage{enumitem}
\usepackage{xparse}

\newif\ifshowcomments
\showcommentsfalse

\ifshowcomments
  \NewDocumentCommand{\newadd}{+m}{{\color{blue}#1}}
  \newcommand{\lu}[1]{\textcolor{red}{[Lu: #1]}}
\else
  \newcommand{\newadd}[1]{#1}   
  \newcommand{\lu}[1]{}         
\fi

\def\BibTeX{{\rm B\kern-.05em{\sc i\kern-.025em b}\kern-.08em
    T\kern-.1667em\lower.7ex\hbox{E}\kern-.125emX}}
\begin{document}

\title{DemoShapley: Valuation of Demonstrations for In-Context Learning}

\author{\IEEEauthorblockN{1\textsuperscript{st} Shan Xie}
\IEEEauthorblockA{\textit{Department of Mathematics and Computer Science} \\
\textit{Eindhoven University of Technology}\\
s.xie1@student.tue.nl}
\and
\IEEEauthorblockN{2\textsuperscript{nd} Man Luo}
\IEEEauthorblockA{\textit{Intel Lab} \\
mluo26@asu.edu}
\and
\IEEEauthorblockN{3\textsuperscript{rd} Chadly Daniel Stern}
\IEEEauthorblockA{\textit{Department of Psychology} \\
\textit{University of Illinois Urbana-Champaign}\\
chadly@illinois.edu}
\and
\IEEEauthorblockN{4\textsuperscript{th} Mengnan Du}
\IEEEauthorblockA{\textit{Department of Data Science} \\
\textit{New Jersey Institute of Technology}\\
mengnan.du@njit.edu}
\and
\IEEEauthorblockN{5\textsuperscript{th} Cheng Lu}
\IEEEauthorblockA{\textit{Department of Computer Science} \\
\textit{University of Illinois Chicago}\\
lucheng@uic.edu}
}


\maketitle

\begin{abstract}
Large language models (LLMs) using in-context learning (ICL) excel in many tasks without task-specific fine-tuning. However, demonstration selection and ordering greatly impact ICL effectiveness. 
 Focus on this issue, we propose \textit{DemoShapley}, a Shapley-value based method that evaluates each demonstration’s contribution by measuring its marginal effect across different prompt permutations. To further account for ICL’s limited context windows and frequent low-shot settings, we introduce \textit{Beta-DemoShapley}, a weighted extension that emphasizes the influence of smaller prompt sizes. Experiments on multiple benchmarks show that DemoShapley consistently outperforms existing influence-based selection strategies, while Beta-DemoShapley further improves performance in low-shot scenarios. Both methods also detect mislabeled data, enhance generalization to out-of-distribution tasks, and reduce demographic bias. Together, they provide a unified and robust framework for demonstration valuation in ICL.


\end{abstract}


\section{Introduction}

Large language models (LLMs) can perform new tasks through in-context learning (ICL), where a few input–label pairs (demonstrations) are included in the prompt instead of fine-tuning the model \citep{brown2020language}. ICL has shown strong performance across diverse tasks with low complexity and reduced resource requirements \citep{smith2022using, touvron2023llama}. However, despite its efficiency, ICL often produces unstable outcomes: accuracy varies widely depending on which demonstrations are chosen, in what order they appear, and how they are formatted \citep{Rubin2021Learning, Li2023, luo2023dr, luo2024context, Chen2022}. Selecting high-quality demonstrations can significantly improve predictions, while noisy or poorly chosen examples may degrade performance  \cite{min2022rethinking}. This sensitivity limits the robustness of ICL in practice.

\lu{we need one paragraph to describe how prior work addresses this problem, and what their limitations are. }

\newadd{
Prior work has explored several strategies for demonstration selection in ICL, such as similarity-based retrieval \citep{liu2021makes, Rubin2021Learning}, diversity-based methods \citep{levy2022diverse}, and influence-based approaches \citep{chang2022careful, nguyen2023context}. 
These methods bring improvements but also have some their own limitations. 
Most methods evaluate demonstrations under a fixed prompt size and order, ignoring the fact that ICL is sequence-sensitive: the same example can help or hurt depending on its position and context. 
Similarity and diversity optimize indirect measures like embedding distance or coverage, which may not align with the true task metric.
Influence-based methods measure the effect of demonstrations on model accuracy, but existing approaches usually compute influence under a fixed number of demonstrations. This ignores how the same example may contribute differently across prompt sizes, which is especially important in ICL where context length is limited.
Some other approaches also rely on gradients or retraining, which is not feasible for closed-source LLMs and does not match the inference-only nature of ICL. 
Finally, existing strategies rarely highlight the impact of small subsets, even though short prompts are the common case due to context limits. 
These limitations motivate our work on DemoShapley and Beta-DemoShapley, which fairly estimate each example’s contribution by averaging over prompt permutations and giving more weight to small-cardinality settings.
}

To tackle this challenge, we introduce two complementary Shapley-value based algorithms for demonstration valuation in ICL. The first, DemoShapley, directly extends Data Shapley \citep{ghorbani2019data} to ICL by quantifying the marginal contribution of each demonstration across multiple prompt permutations, thereby accounting for order sensitivity. The second, Beta-DemoShapley, is inspired by Beta Shapley \citep{kwon2021beta} and incorporates a weighted formulation that emphasizes smaller subsets, reflecting the short context windows and computational limits typical of ICL. Together, these methods provide a principled framework for fairly and effectively evaluating the utility of demonstrations.

Data Shapley is grounded in cooperative game theory \citep{shapley1953value}, where the Shapley value quantifies each participant’s average marginal contribution across all coalitions. Previous work has successfully applied this concept to data valuation in traditional machine learning \citep{gardner2020evaluating, kwon2021beta, jia2019efficient}, but these approaches have not addressed the sequence-dependent effects unique to ICL. By adapting Shapley-based estimation to prompts, DemoShapley and Beta-DemoShapley capture how examples influence performance under both general and low-shot conditions.

\begin{figure*}
    \begin{center}
\includegraphics[width=\textwidth]{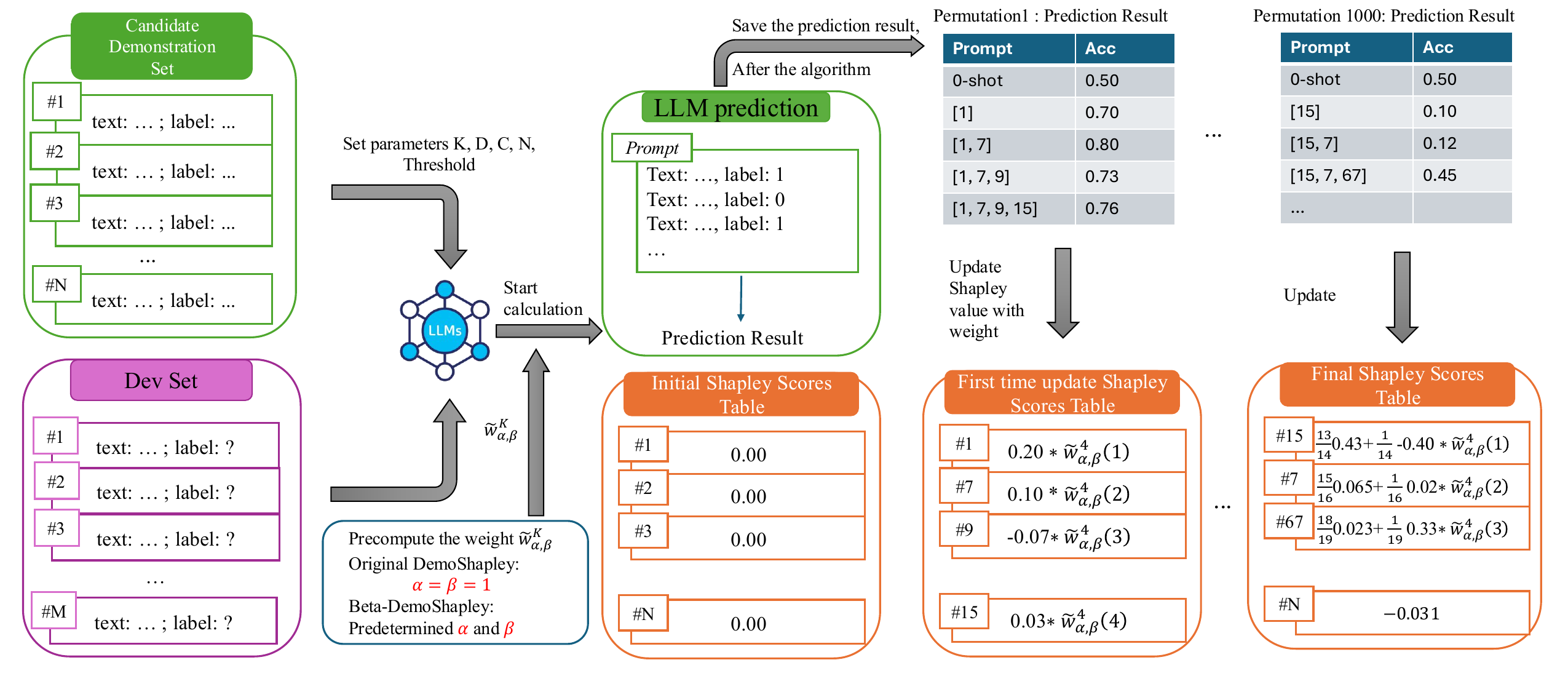}
    \end{center}
    \caption{
    We begin by selecting a candidate demonstration set and a development dataset, then define hyperparameters $K, D, C, N$ and threshold $\mu$. Starting from zero-shot learning, examples are added sequentially, with their DemoShapley values updated per prompt. Beta-DemoShapley differs by applying a pre-computed weight to the marginal contribution during updates. The algorithm iterates multiple times to ensure all candidate examples are thoroughly evaluated before concluding.}
    \label{fig:proposed-framework}
\end{figure*}

In practice, both algorithms assess each demonstration by running it through different prompt permutations, so that the effect of ordering is averaged out. This allows us to see whether an example consistently helps or harms model predictions, rather than judging it from a single prompt. Fig.~\ref{fig:proposed-framework} illustrates this process and shows how the framework highlights demonstrations with positive or negative impact.


Through comprehensive empirical validation, we show that our algorithms not only effectively select demonstrations that enhance model performance but also significantly improve the model's generalization capabilities, particularly in out-of-distribution (OOD) tasks. This is critical for machine learning models deployed in real-world settings, where they often encounter data that differs from the training distribution. Further, our proposed algorithms offer advantages in identifying demonstrations that may be affected by label noise, thus preventing them from adversely impacting the model's performance in advance. Lastly, we show that selecting high-value data also enhances fairness in LLM performance. Taken together, these results show that Shapley-based valuation makes ICL more reliable and robust in practice.

\section{Related Work}

\subsection{Data Valuation}

Data valuation aims to quantify the importance of individual data points in machine learning. For a detailed overview, we refer readers to \citet{sim2022data}. Existing approaches can be grouped into three main families: Leave-One-Out (LOO), cooperative game theory (CGT)-based, and desiderata-driven methods.
LOO-based methods, introduced by \citet{cook1980characterizations}, evaluate a data point \( D_k \) by measuring changes in model performance when \( D_k \) is removed from the dataset. 
CGT-based methods, inspired by cooperative game theory, estimate the marginal contribution of \( D_k \) across different coalitions, with the Shapley value \citep{shapley1953value} and Banzhaf index \citep{banzhaf1964weighted} as canonical examples. Such contributions can be either positive or negative.

Desiderata-driven methods adapt Shapley principles to meet practical requirements, leading to variants such as Data Shapley \citep{ghorbani2019data}, Distributional Shapley \citep{ghorbani2020distributional}, Beta Shapley \citep{kwon2021beta}, and KNN-Shapley \citep{jia2019efficient}. Other strategies include DVRL \citep{yoon2020data}, which formulates valuation as a reinforcement learning problem, and influence-based approaches that track gradient descent dynamics \citep{pruthi2020estimating}. These methods trade off theoretical guarantees with computational efficiency or specific desiderata such as robustness and scalability.

\subsection{ICL Demonstration Selection} 

\citet{chang2022careful} found that selecting training samples based on their impact on model accuracy stabilizes ICL outcomes compared to random selection. Similarly, \citet{nguyen2023context} developed a method to assess and select impactful examples, boosting performance on SuperGLUE tasks. In retrieval-based ICL, demonstration selection often prioritizes similarity and diversity. Similarity-based methods choose examples aligned with the query using linguistic or structural features. \citet{liu2021makes} showed that semantically closer demonstrations improve GPT-3 ICL performance with kNN retrieval, while \citet{Rubin2021Learning} and \citet{cheng2023uprise} proposed retrieval methods leveraging language model cues and cross-domain datasets, respectively.  

Beyond similarity, diverse demonstrations help prevent redundancy, capture varied viewpoints, and improve test coverage \citep{levy2022diverse}. Additionally, demonstrations with higher complexity, such as longer queries or more reasoning steps, can enhance ICL performance \citep{fu2022complexity}.  

Recent work has explored learning-based selection. \citet{liu2024unraveling} identify two key signals—task-agnostic input similarity and task-specific output similarity—and propose lightweight methods (MLSM and TTF) that improve ICL without requiring large auxiliary datasets. In parallel, \citet{zhang2025selectingdemonstrationsmanyshotincontext} introduce a gradient-matching strategy for many-shot ICL, aligning gradients of selected examples with those of the full training set and outperforming random selection across 4–128 shot settings.  

Other studies optimize demonstration choice more directly. \citet{zhang2025learning} present the GenICL framework, which applies generative preference learning with LLM feedback and achieves state-of-the-art results across diverse benchmarks. \citet{zhang2025linear} propose a linear-time method based on gradient estimation in the input embedding space, which improves over embedding-similarity baselines by up to 11\% while being substantially faster. Finally, \citet{kato2025affinity} combines affinity (query similarity) and diversity into a unified metric based on internal representations, showing strong correlation with ICL performance. These works represent a move beyond simple similarity heuristics, aiming instead at more systematic and scalable strategies for demonstration selection. 
\lu{need a paragraph for our unique contributions compared to related work}

\newadd{Compared with existing work, our approach introduces several unique advantages and contributions.
First, our two DemoShapley methods account for the order sensitivity of ICL by averaging contributions across multiple prompt permutations, while prior methods typically evaluate demonstrations under a fixed order. 
Second, Beta-DemoShapley incorporates cardinality-aware weights, giving greater importance to smaller subsets, which better matches the short-context regime of ICL. 
Third, unlike some other methods that require gradient access or fine-tuning, our framework relies only on the inference, making it applicable to closed-source LLMs. 
Finally, by extending Data Shapley \citep{ghorbani2019data}, our method preserves its fairness guarantees rooted in cooperative game theory: each demonstration is valued according to its average marginal contribution across contexts, ensuring that contributions are measured consistently and without bias. 
Taken together, these properties include order awareness, cardinality weighting aligned with ICL, applicability to closed-source LLMs, and fairness guarantees. 
These features make our DemoShapley and its Beta extension a principled and practical framework for demonstration valuation.}
\section{Demonstration Valuation for ICL}
\subsection{Preliminaries}
\subsubsection{Data Shapley}



Data Shapley \citep{ghorbani2019data} is a method for fair data valuation in machine learning. It assigns each data point a value that reflects its contribution to the performance of a learning algorithm. Formally, given a dataset $D$, a learning algorithm $A$, and a performance metric $V$, the Shapley value of the $i$-th data point is defined as
\begin{equation}
    \phi_i = \sum_{S \subseteq D \setminus \{i\}} \binom{n-1}{|S|}^{-1} \big[V(S \cup \{i\}) - V(S)\big],
\end{equation}
where $n$ is the total number of data points, $S$ is a subset of $D$ excluding the $i$-th point, and $V(S)$ denotes the performance of the predictor trained on $S$. The binomial coefficient $\binom{n-1}{|S|}$ normalizes the marginal contribution over all subsets of size $|S|$.

\subsubsection{Beta Shapley}
A limitation of Data Shapley is that when the subset size $\lvert S \rvert$ becomes large, the marginal contribution of individual points tends to vanish. As a result, its uniform weighting scheme may fail to capture meaningful differences between examples. To address this issue, \textit{Beta Shapley} \citep{kwon2021beta} introduces a weighting function based on the Beta distribution, which emphasizes smaller subsets and thus better reflects the contributions of individual data points.


\subsection{From Data Shapley to (Beta-)DemoShapley}


In ICL, demonstrations play a role similar to training data in standard machine learning. \textit{DemoShapley} extends Data Shapley by estimating the marginal contribution of each demonstration: demonstrations are added to the prompt in different orders, and the resulting change in model performance reflects their importance. \textit{Beta-DemoShapley}, inspired by Beta Shapley, modifies this framework by giving higher weights to smaller subsets. This adjustment is well aligned with ICL, where prompt length is limited and each demonstration can have a substantial impact. By emphasizing contributions in low-cardinality settings, Beta-DemoShapley captures the characteristics of ICL more faithfully.  

Although DemoShapley avoids retraining and only relies on model inference, exact computation is still expensive because it requires evaluating an exponential number of prompt permutations. To make the approach practical, we adopt the \textbf{Truncated Monte Carlo (TMC)} method~\citep{maleki2013bounding, ghorbani2019data, kwon2021beta}, which approximates Shapley values through random sampling of subsets, greatly reducing the computational cost of (Beta-)DemoShapley.

\subsubsection{Weight Computation for (Beta-)DemoShapley}

A limitation of Data Shapley is that its uniform weighting scheme cannot distinguish the relative importance of subsets of different sizes. In ICL, this issue becomes particularly relevant: the context window is short and prompts typically contain only a limited number of demonstrations, so the marginal contribution of an example within a small subset is much more informative than its contribution within a large one. To reflect this property, (Beta-)DemoShapley applies predetermined weights that emphasize smaller subset cardinalities. This adjustment helps the method better align with the constraints of ICL, where each demonstration can substantially affect the predictions of the model.  

Formally, we adopt the Beta distribution with positive hyperparameters $\alpha$ and $\beta$ to define the weight $w$:  
\begin{equation}
\begin{aligned}
w_{\alpha, \beta}^{(n)}(j) 
& := n \int_0^1 t^{j-1}(1-t)^{n-j} 
   \frac{t^{\beta-1}(1-t)^{\alpha-1}}{\operatorname{Beta}(\alpha, \beta)} \, dt \\
& = n \cdot \frac{\operatorname{Beta}(j+\beta-1, n-j+\alpha)}{\operatorname{Beta}(\alpha, \beta)},
\end{aligned}
\end{equation}
where $\operatorname{Beta}(\alpha, \beta) = \frac{\Gamma(\alpha) \Gamma(\beta)}{\Gamma(\alpha + \beta)}$ is the Beta function, $\Gamma(\cdot)$ is the Gamma function, $n$ is the dataset size, and $j$ is the cardinality of the candidate subset. This expression can be simplified to
\begin{equation}
w_{\alpha, \beta}^{(n)}(j)
= n \cdot \frac{\prod_{k=1}^{j-1}(\beta+k-1)\prod_{k=1}^{n-j}(\alpha+k-1)}
        {\prod_{k=1}^{n-1}(\alpha+\beta+k-1)}.
\end{equation}

Different settings of $(\alpha,\beta)$ cover various aspects of demonstration contribution.  
\begin{itemize}[leftmargin=*]
    \item When $\alpha > 1$ and $\beta=1$, the weighting scheme places greater emphasis on \textit{smaller subsets}. In ICL, this corresponds to focusing on prompts that contain only a few demonstrations, where each example tends to have a strong influence. At the same time, the effect of noisy signals from very large subsets is reduced.  

    \item When $\alpha=1$ and $\beta>1$, the emphasis shifts toward \textit{larger subsets}. Here, the importance of a demonstration is judged in the context of many other examples, which is more suitable when long prompts are available.  

    \item The special case $\alpha=\beta=1$ recovers the uniform scheme of Data Shapley. In this setting, all subset sizes are weighted equally, and the normalized weight satisfies $\Tilde{w}_{\alpha,\beta}^{(n)}(j)=1$ for all $j \in [n]$, giving the original DemoShapley algorithm.  
\end{itemize}
Once the weights $\Tilde{w}_{\alpha,\beta}^{(n)}(j)$ are calculated for a chosen $(\alpha,\beta)$ and the size of the data set $n$, they can be reused throughout the estimation process. This one-time calculation is negligible compared to the cost of LLM inference, so the weighting scheme does not increase the runtime complexity of Beta-DemoShapley.
\subsubsection{A Unified Framework}

We propose a unified framework that accommodates both DemoShapley and Beta-DemoShapley within the same formulation. To adapt TMC-Shapley to ICL, we modify the calculation to reflect that prompts, rather than full training datasets, serve as the evaluation unit. With in this framework, the Shapley value of demonstration $i$ can be expressed as a weighted mean of its marginal contributions across sampled subsets:


\begin{equation}
    \phi_i = \frac{1}{K}\sum_{j=1}^{K}\frac{1}{|D_{j}^{\backslash i}|}
    \sum_{S \in D_{j}^{\backslash i}} \Tilde{w}_{\alpha, \beta}^{(n)}(j)\big[V(S \cup \{i\}) - V(S)\big],
\end{equation}
where $D_{j}^{\backslash i}$ denotes subsets of size $j$ that exclude $i$, and $\Tilde{w}_{\alpha,\beta}^{(n)}(j)$ provides the normalized weight for cardinality $j$. Setting $\alpha=\beta=1$ recovers DemoShapley; other values yield Beta-DemoShapley.

The process is described as follows:
\newline

\begin{enumerate}[leftmargin=*]
\item \textbf{Weight Computation}: Given $(\alpha,\beta)$ and $n$, compute $\Tilde{w}^{(n)}_{\alpha,\beta}(j)$ for all $j=1,\dots,K$ once.
Setting $\alpha=\beta=1$ recovers DemoShapley (uniform weights); other settings give Beta-DemoShapley.
\vspace{1em}
\item \textbf{Random Sample and Permutation}: 
From the candidate pool $D$, uniformly sample $K$ demonstrations to form a subset $D_K$ with $|D_K|=K$. 
Then draw a random permutation $\Pi=\{\pi_1,\pi_2,\ldots,\pi_K\}$ of the selected demonstrations. 
These demonstrations will be inserted into the prompt one by one following $\Pi$. 
The initial prompt $P_0$ contains only the task instructions (zero-shot case), and demonstrations are added sequentially in later steps.
\vspace{1em}
\item \textbf{Performance Metric Calculation}: Evaluate the LLM $f$ with prompt $P$ and compute the metric $V(P,f)$. As a baseline, record the zero-shot performance $V(\emptyset,f)$, where the prompt consists solely of the task instruction, needs to be calculated.
\vspace{1em}
\item \textbf{Sequential Demonstration Addition}: Add demonstrations sequentially following $\Pi$, updating the performance score after each addition. The same development dataset is used throughout to ensure comparability.
\vspace{1em}
\item \textbf{Shapley Value Update}: If the performance change \(v'\) with a newly added demonstration exceeds a certain threshold, update the DemoShapley value \(\phi_n\) for the newly added demonstration \(\pi_n\) using the equation:
\begin{equation}
        \phi_{\pi^{t_c}[c]} \leftarrow \frac{t_c-1}{t_c} \phi_{\pi^{t_c-1}[c]}+\Tilde{w}^{(n)}_{\alpha,\beta}\left(j\right)\frac{1}{t_c}\left(v^{'}\right),
\end{equation}
where \(t_c\) denotes the number of iterations for the \(c\)-th demonstration. For Beta-DemoShapley, the precomputed weight $\Tilde{w}^{(n)}_{\alpha,\beta}(j)$ is applied to the performance change.
\vspace{1em}
\item \textbf{Repetition for Convergence}: The process terminates under two conditions:
\textbf{(1) Max Iterations Times}: a maximum of $N$ iterations is reached, or  
\textbf{(2) MC Convergence Check}: our proposed algorithm is based on MC method which guarantees to converge to be true value if we repeat the sampling procedure. We use the Gelman-Rubin statistic \cite{gelman1995bayesian}, a Monte Carlo (MC) convergence diagnostic, to stop sampling when the performance increment becomes negligible.
\end{enumerate}
These steps yield an efficient and stable estimator of our Shapley-based method's values. Our adapted framework quantifies each demonstration’s contribution under ICL constraints, enabling more reliable selection for improved model performance. The complete pseudocode is provided in Algorithm format ~\ref{alg:cap}.
\begin{algorithm}[H]
\small
\caption{Efficient Computation Algorithm for DemoShapley and Beta($\alpha,\beta$)-DemoShapley}
\label{alg:cap}
\textbf{Input} Number of demonstrations \( K \), Candidate Set \( \mathcal{C} = \{c_1, \ldots, c_n\} \), Dev Set \( \mathcal{D} \), Termination Threshold \( \rho \) (set to 1.005), Hyperparameters \( \alpha, \beta \), Max Iterations \( N \), Validation Method \( V \), LLM \( f \), Threshold \( \mu \), Instruction Prompt \( \mathcal{P_I} \).

\textbf{Output} Shapley Scores Table $\Phi$ with size of $\left| \mathcal{C} \right|$ for candidate demonstration set $\mathcal{C}$
\begin{algorithmic}[1]
\State Intialize $\hat{\rho}=2\rho, B=1, \Phi() $
\State zero\_shot\_acc $\gets V(\mathcal{P_I},f, D)$
\State Compute $\Tilde{w}^{(n)}_{\alpha,\beta} \leftarrow \left(^{n-1}_{j-1}\right)w^{(n)}_{\alpha,\beta}(j)$  for all $j \in \left[ n \right]$ ($\alpha$ and $\beta$ set to 1 when running Original DemoShapley)
\For{$i \gets 1$ to $\left[ n \right]$}
\State $t_i \gets 0$   \Comment{Iteration times for $i_{th}$ datum}
\EndFor
\For{$i \gets 1$ to $N$}
\State Select $K$ datum from Candidate Demonstration Set $\mathcal{C}$
\State Generate a random permutation $\Pi$ with $K$ datum selected.
\State last\textunderscore acc $\gets$ zero\_shot\_acc
\State Prompt $\mathcal{P} \gets\mathcal{P_I} $
\For{$j \gets 1$ to $K$}
\State $\mathcal{P} \gets \mathcal{P}+\pi_j$
\State acc$\gets V(\mathcal{P}, f)$
\If{ $\lvert$acc$-$ last\_acc $\rvert\geq \mu$}
    \State $t_c \gets t_c +1$
    \State $v^{'} \gets $acc$-$last\_acc 
    \State $\phi_{\pi^{t_c}[c]} \leftarrow \frac{t_c-1}{t_c} \phi_{\pi^{t_c-1}[c]}+\frac{1}{t_c}\Tilde{w}^{(n)}_{\alpha,\beta}\left(j\right)\left(v^{'}\right)$
\EndIf
\EndFor
\State Update the Gelman-Rubin statistic $\hat{\rho}$
\EndFor
\end{algorithmic}
\end{algorithm}
\section{Experiments \& Applications}
In this section, we evaluate DemoShapley and Beta-DemoShapley through four sets of experiments: (1) predictive performance, (2) detecting label noise, (3) generalization on OOD tasks, and (4) LLMs fairness analysis.

\subsection{Experimental Settings}
\noindent\textbf{LLMs:}
Our evaluation framework includes four primary models: two GPT-structured models—ChatGPT-3.5-Turbo \citep{OpenAi} and GPT-J-6B \citep{wang2021gpt}—and two non-GPT models—Mistral-7B-v0.3 \citep{jiang2023mistral} and Llama3-8B \citep{llama3modelcard}.

\noindent\textbf{Datasets:}
\newadd{
We consider six datasets covering a range of tasks.  
Toxi-text-3M classifies texts as toxic or non-toxic, and we focus on its English subset \citep{fredzhang7toxi}.  
The Adult dataset predicts income levels from demographic and socioeconomic attributes \citep{misc_adult_2}.  
The Emotion dataset labels Twitter messages with six emotions \citep{saravia2018carer}.  
BoolQ is a yes/no question-answering benchmark, and BoolQ Contrast tests model adaptability to distribution shifts \citep{clark2019boolq, gardner2020evaluating}.  
SST-2 analyzes sentiment in movie reviews \citep{socher2013recursive}, and FinancialPhraseBank-v1.0 contains sentences from financial news articles \citep{malo2014good}.  

These datasets are used in our four experimental settings.  
For \textit{Prediction Efficacy}, we use Adult, Toxi-text-3M, and Emotion.  
For \textit{OOD Generalization}, we adopt BoolQ, BoolQ Contrast, SST-2, and FinancialPhraseBank.  
For \textit{Robustness to Label Noise}, we construct noisy variants of Adult and Toxi-text-3M by flipping 10\% of the labels with manual verification.  
For \textit{Algorithmic Fairness}, we again use Adult, following standard practice with sensitive attributes.
}

\noindent\textbf{Implementation Details:} 
For DemoShapley, we use the uniform weighting scheme with $\alpha=\beta=1$. 
For Beta-DemoShapley, we set $\alpha=4$ and $\beta=1$ to emphasize smaller subset sizes, which better reflects the short-context regime of ICL. 
Unless otherwise noted, we use temperature $=0$ for deterministic evaluation and repeat each experiment three times, reporting the mean score to reduce variance. Other hyperparameters vary and are indicated in different the experiment descriptions..

\noindent\textbf{Baseline Methods:}
We compare DemoShapley and Beta-DemoShapley against four baseline demonstration selection methods: CondAcc \citep{chang2022data} and Influence \citep{nguyen2023context}, which represent state-of-the-art influence-based strategies, as well as Leave-One-Out (LOO)~\citep{cook1977detection} and Random selection.

\begin{itemize}[leftmargin=*]\setlength\itemsep{-0em}
\item \textbf{CondAcc} selects demonstrations that enhance and stabilize ICL performance. The CondAcc score for a demonstration $x_i$ with LLM $f$ is computed as: $\boldsymbol{s}_{c a}(i)=\mathbb{E}_{\mathcal{Z} \sim \mathcal{D}_{\mathrm{ICL}}}\left[f(\mathcal{Z}) \mid x_i \in \mathcal{Z}\right]$, where $\mathcal{Z}$ denotes the prompt. 
\item \textbf{Influence} measures the impact of each example by comparing the average performance of prompts with and without $x_i$. The calculation is given as $\mathcal{I}\left(i\right)=\frac{1}{N_i} \sum_{\mathcal{Z}: x_i \in \mathcal{Z}} f\left(\mathcal{Z}\right)-\frac{1}{M-N_i} \sum_{\mathcal{Z}: x_i \notin \mathcal{Z}} f\left(\mathcal{Z}\right)$ where $N_i$ is the number of prompts containing $x_i$ and $M$ is the total number of prompts.

\item \textbf{LOO} evaluates a data point’s significance by measuring the change in model performance when being removed. For ICL, we assess the performance difference between a complete prompt and one excluding $x_i$ to quantify its contribution.
\item \textbf{Random} selection randomly samples $K$ examples from the candidate demonstration pool for inference in ICL.
\end{itemize}

\subsection{The Prediction Efficacy of DemoShapley}

We design two experiments to showcase the predictive performance of DemoShapley compared to other baselines. \lu{what are the evaluation metrics? Since all 4 experiments have different settings, please make sure you introduce the experimental setting in each experiment.}

\newadd{\textbf{Experimental setting.}  
We conduct experiments on the Adult, Toxi-text-3M, and Emotion datasets, covering both binary and multi-class classification tasks.  
From each dataset, we randomly sample 100 instances as the candidate demonstration pool.  
Evaluation is performed on the corresponding test sets, using accuracy for binary tasks (Adult, Toxi-text-3M, SST-2) and macro-F1 for multi-class tasks (Emotion).  
We report results for GPT-structured models (GPT-3.5-Turbo, GPT-J-6B) and non-GPT models (Llama3-8B, Mistral-7B).  
The zero-shot baseline uses only task instructions without demonstrations. } 

\textit{Add In-context Learning Examples based on Demonstration Values.} 
We start with a zero-shot setting and apply two selection strategies. The first strategy incrementally adds examples with the highest Demonstration Shapley values, enriching the prompt with valuable demonstrations. The second strategy also begins from a zero-shot baseline but instead adds examples with the lowest values, introducing demonstrations that are theoretically less beneficial or even detrimental.

\textit{Remove In-context Learning Examples based on Demonstration Values.} 
We begin with 10 randomly selected demonstrations and conduct two removal experiments. The first removes the top five highest-valued examples to assess their impact on prediction performance. The second removes the five lowest-valued examples to evaluate how eliminating less valuable demonstrations affects performance.

\begin{figure*}[htbp]
    \centering
    \includegraphics[width=0.85\textwidth]{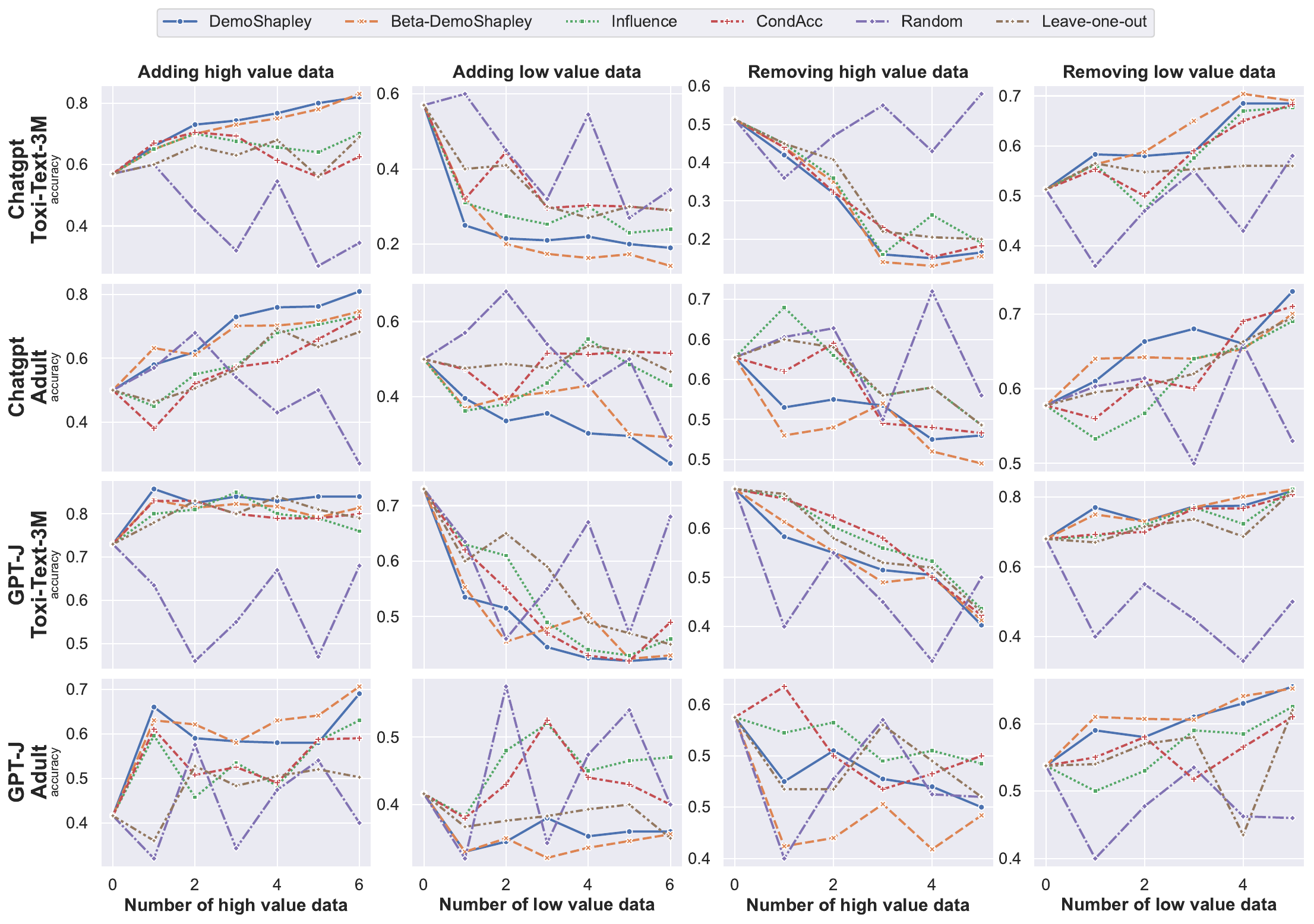}
    \caption{Effect of demonstration values on predictive performance across Toxi-text-3M and Adult datasets using ChatGPT-3.5-Turbo and GPT-J-6B. 
Columns show adding/removing high- or low-value demonstrations. 
DemoShapley and Beta-DemoShapley align well with actual performance: high-value samples improve accuracy, while low-value samples tend to harm it.}
    \label{exp1}
    \vspace{-3mm}
\end{figure*}

\textbf{Results} We have the following observations based on the results for GPT models in Figure \ref{exp1} and results for non-GPT architecture LLMs in Table \ref{Additional_result}.

\textit{Observation 1: High DemoShapley Value Demonstrations Tend to Improve Performance, While Low-Value Ones Are More Likely to Be Detrimental.} 
The first column of Figure \ref{exp1} shows that adding high-value demonstrations generally improves performance more than other methods. 
Notably, while high DemoShapley value demonstrations are generally beneficial, adding too many can sometimes harm performance, even if their values are positive.
In contrast, the second column and Table \ref{Additional_result} indicate that adding low-value demonstrations often leads to performance decline, a trend more pronounced with the our methods. This suggests a strong correlation between DemoShapley values and their actual impact in ICL, though their influence is not strictly linear. The trend remains consistent across different model architectures, including Llama3 and Mistral, reinforcing that data quality plays a crucial role in model performance, regardless of the model structure.
Beta-DemoShapley performs slightly better than DemoShapley, particularly when the number of demonstrations in the prompt is small, highlighting its effectiveness in emphasizing performance at lower cardinalities. 

\textit{Observation 2: Eliminating demonstrations with high DemoShapley values leads to reduced performance, whereas removing those with low values enhances model performance. }
The third column of Figure~\ref{exp1} shows that removing the highest DemoShapley value demonstrations from a set of 10 significantly reduces performance, with the sharpest drop occurring after removing the top five. Conversely, the fourth column demonstrates that eliminating negative-value demonstrations rapidly improves performance. Notably, removing the highest net-value demonstrations causes the most substantial performance change, emphasizing their strong influence on ICL effectiveness. This highlights the importance of retaining high-value demonstrations while filtering out low-value ones for optimal results.

\begin{table*}[]
\centering
\caption{Results for Llama3 and Mistral models across different datasets under the same experimental settings are presented in the experiment section. We compared our methods with other benchmarks that assign scores to each datum. The results indicate that our approach better captures the intrinsic relationship between demonstrations and their actual impact on model performance.}
\vspace{-1.0em}
\begin{adjustbox}{width=.85\textwidth}
\begin{tabular}{rcccccccccc}
\toprule[1.5pt]
\multicolumn{1}{c}{} & \multicolumn{5}{c}{\begin{tabular}[c]{@{}c@{}}Adding High Value Data (Accuracy \textbf{\%})$\uparrow$\end{tabular}}  & \multicolumn{5}{c}{\begin{tabular}[c]{@{}c@{}}Adding Low Value Data (Accuracy \textbf{\%})$\downarrow$\end{tabular}}\\
\cmidrule(lr){2-6} 
\cmidrule(lr){7-11}
\multicolumn{1}{c}{} Methods & DemoShapley  & Beta-DemoShapley   & CondAcc    & Influence  & LOO  & DemoShapley & Beta-DemoShapley & CondAcc   & Influence  & LOO  \\
\midrule[1pt]
Llama3 (Toxi-Text-3M)  & 86.7 & \textbf{87.2} & 85.0 & 84.3 & 80.3 & \textbf{21.0} & 22.3 & 27.7 & 33.0 & 40.7\\
Llama3 (Adult)         & 79.7 & \textbf{80.1} &74.5 & 74.7 & 73.7 & 47.5 & \textbf{45.3} & 58.7 & 52.0 & 57.6 \\
Llama3 (Emotion)  & \textbf{47.7} & 46.8 & 43.0 & 43.2 & 44.3 & 23.5& 20.2 &21.5 &23.0& \textbf{19.3} \\
Mistral (Toxi-Text-3M) & 82.1 & \textbf{83.1} &80.3 & 76.3 & 65.3 & \textbf{40.7}& 41.1 & 46.7 & 40.3 & 44.6 \\
Mistral (Adult)       & 73.5 & 73.7 &\textbf{74.7} & 71.3 & 67.1& 49.7 & \textbf{49.5} & 51.2 & 52.2 & 57.0 \\
Mistral (Emotion)      & \textbf{44.3}& 44.1 & 43.0 & 44.5 & 40.2& 15.0& 15.3 & \textbf{14.2} & 25.0 & 16.1 \\
\midrule[1pt]
\multicolumn{1}{c}{} & \multicolumn{5}{c}{\begin{tabular}[c]{@{}c@{}}Removing High Value Data (Accuracy \textbf{\%})$\downarrow$\end{tabular}}  & \multicolumn{5}{c}{\begin{tabular}[c]{@{}c@{}}Removing Low Value Data (Accuracy \textbf{\%})$\uparrow$\end{tabular}}\\
\cmidrule(lr){2-6} 
\cmidrule(lr){7-11}
Llama3 (Toxi-Text-3M)  & 53.0 & \textbf{52.4} & 52.7 & 53.2 & 54.8 & \textbf{82.3}& 80.3 & 76.7 & 77.4 & 74.6 \\
Llama3 (Adult)         & \textbf{64.5} & 65.3 & 67.5 & 69.5 & 68.6  & 72.3 & 74.1 & 72.0 & 71.5 & \textbf{74.5} \\
Llama3 (Emotion)   & 29.7 & \textbf{25.6} & 31.9 & 32.8 & 30.4 & 45.3& \textbf{46.4} & 40.3 & 39.5 & 41.7 \\
Mistral (Toxi-Text-3M) & 44.3& 43.1 & 45.5 & \textbf{42.7} & 43.8 & 65.7& \textbf{66.3} & 63.2 & 62.3 & 64.8 \\
Mistral (Adult)        & \textbf{60.5}& 61.2 & 61.8 & \textbf{60.5} & 63.5 & 72.3& 72.5  & 70.3 & \textbf{73.0} & 66.5 \\
Mistral (Emotion)  & 24.3& \textbf{23.6} & 29.8 & 26.4 & 29.0 & 38.6& \textbf{39.1} & 37.7 & 37.0 & \textbf{39.1}\\
\bottomrule[1.5pt]
\end{tabular}
\end{adjustbox}

\label{Additional_result}
\end{table*}

\vspace{-0.5em}

\subsection{DemoShapley for OOD Generalization}
\lu{what are the evaluation metrics?}

\newadd{
\textbf{Experimental Settings.}  
We evaluate whether higher-valued demonstrations according to (Beta-)DemoShapley improve model generalization on out-of-distribution (OOD) tasks without fine-tuning. We consider two transfer scenarios:  
(1) BoolQ as the source dataset with BoolQ Contrast as the OOD test set, and  
(2) transferring sentiment analysis from SST-2 to FinancialPhraseBank-v1.0.  
We use GPT-J-6B for evaluation, since ChatGPT-3.5-Turbo shows minimal prediction variation across permutations, likely because both datasets are included in its pre-training. \newadd{Performance is measured by classification accuracy on the OOD test sets.}  
}
Our results are presented in Table \ref{ood}. 

\textit{Observations:} Our results show that our method outperforms baselines on OOD tasks under the same experimental conditions. Beta-DemoShapley performs better with fewer demonstrations, indicating that assigning higher weights to smaller cardinalities is effective. The performance gap is especially evident on FinancialPhraseBank-v1.0, highlighting the advantages of our selection strategy. These findings confirm that our method not only excels on in-distribution testsets but also enhances the model’s ability to interpret nuanced prompt information, resulting in better generalization in ICL scenarios.

\begin{table}[H]
\caption{\newadd{Accuracy on BoolQ Contrast and FinancialPhraseBank-v1. 
Numbers in parentheses denote zero-shot baselines (0.502 for BoolQ Contrast, 0.553 for FinancialPhraseBank). 
Prompts consist of demonstrations selected by different evaluation algorithms.}}
\centering
\scalebox{0.95}{
\begin{tabular}{ccccc}
\toprule
\multirow{2}{*}{GPT-J-6B OOD Tasks} & \multicolumn{2}{c}{\textbf{BoolQ Cst.(0.502)}} & \multicolumn{2}{c}{\textbf{Financial (0.553)}} \\ 
\cmidrule(lr){2-3} \cmidrule(lr){4-5} 
 & 16-shots & 32-shots & 16-shots & 32-shots \\
\midrule
DemoShapley & 0.568  & \textbf{0.582} & 0.652 &  \textbf{0.674}  \\
Beta-DemoShapley & \textbf{0.574} & 0.579 &\textbf{0.663}& 0.655\\
CondAcc & 0.554 & 0.561 & 0.614 & 0.632 \\
Influence & 0.556 & 0.553 & 0.622 & 0.620 \\ 
LOO & 0.535    & 0.564    & 0.594 & 0.582 \\
Random & 0.562 & 0.531 & 0.532 & 0.568 \\
\bottomrule
\end{tabular}}
\label{ood}
\end{table}

\subsection{DemoShapley Robust to Label Noise}
\lu{what are the evaluation metrics?}
\newadd{
Label noise is a common challenge in crowd-sourced data sets, where misannotations can significantly reduce the reliability of the model. Building on our motivation that Shapley-based methods should naturally downweight mislabeled examples, we design experiments to test whether (Beta-)DemoShapley can serve as an effective tool for noise detection.  

\textbf{Experimental Settings.}  
We evaluate whether our proposed methods can identify mislabeled demonstrations. Following \citet{ghorbani2019data}, who applied Data Shapley for noisy-label detection, we expect similar benefits in ICL. We use the Adult dataset \citep{misc_adult_2} and the English subset of Toxi-text-3M \citep{fredzhang7toxi}. For each dataset, we construct a demonstration pool of 100 randomly sampled examples, and introduce label noise by flipping 10\% of the labels with manual verification to ensure realistic mislabeling. To measure detection performance, we rank demonstrations by their estimated values and compute how many flipped labels appear among the lowest-valued 10 and 20 examples (bottom-10 / bottom-20 analysis). This metric reflects whether mislabeled examples consistently receive low Shapley values.}

\textbf{Results}
Our results are shown in Table \ref{flippedlabeltable}. We have observations below:

\textit{Obvervation 1:} Our method demonstrates superior sensitivity in detecting label-flipped samples in toxic classification tasks across both ChatGPT-3.5-Turbo and GPT-J-6B, outperforming other approaches. However, its performance on the Adult dataset is less robust, likely because label flipping in toxic classification strongly contradicts pre-trained semantics, whereas in the Adult dataset, flipped labels deviate from the original distribution without semantic contradiction. This trend is also observed with LOO, while CondAcc and Influence perform better on the Adult dataset, suggesting they adapt more effectively to its label-flipping characteristics.

\textit{Observation 2:} Larger models are more sensitive to label noise than smaller ones, as reflected in Average Rank, which measures the overall impact of flipped-label examples. ChatGPT-3.5-Turbo consistently achieves a better average rank than GPT-J-6B across all datasets. Lower scores indicate a stronger negative impact of flipped labels in ICL, aligning with \citet{wei2023larger}. This suggests that larger models can override semantic priors when exposed to in-context flipped labels, while smaller models depend more on pretrained priors and tend to disregard label noise.
\begin{table}[htbp]
\centering
\caption{\newadd{Label-noise detection results on Toxi-text-3M and Adult datasets using ChatGPT-3.5 and GPT-J-6B. 
We introduce 10\% label flips and rank demonstrations by their evaluation scores under different methods. 
For each method, we report (i) the number of flipped labels appearing in the lowest 10 and 20 ranked examples (``N in bottom 10/20''), 
and (ii) the average rank position of flipped labels (lower is better).}}
\vspace{-1.0em}
\begin{adjustbox}{width=0.5\textwidth}
\begin{tabular}{ccccccc}
\toprule
\multirow{2}{*}{ChatGPT-3.5} & \multicolumn{3}{c}{\textbf{Toxi}}& \multicolumn{3}{c}{\textbf{Adult}} \\
 \cmidrule(lr){2-4} \cmidrule(lr){5-7} 
 & \makecell[c]{N in \\ bottom 10} &\makecell[c]{N in \\ bottom 20}& \makecell[c]{Average \\ Rank} &\makecell[c]{N in \\ bottom 10}& \makecell[c]{N in \\ bottom 20} &   \makecell[c]{Average \\ Rank} \\
\midrule
DemoShapley & 3 & \textbf{7}& \textbf{22.3} & \textbf{2} & 5 & 24.2 \\
Beta-DemoShapley & \textbf{4} & 6& 23.3 & \textbf{2} & 5 & 28.5 \\
CondAcc & 2 & 5 & 25.9 & \textbf{2} & \textbf{6} & \textbf{22.2} \\
Influence & 2 & 5 & 25.9 & \textbf{2} & \textbf{6} &\textbf{22.2} \\
LOO & 3 & 6 & 29.1 & \textbf{2} & 2 & 30.8 \\
\bottomrule
\end{tabular}
\end{adjustbox}
\newline
\vspace*{0.3cm}
\newline
\begin{adjustbox}{width=0.5\textwidth}
\begin{tabular}{ccccccc}
\toprule
\multirow{2}{*}{GPT-J-6B} & \multicolumn{3}{c}{\textbf{Toxi}}& \multicolumn{3}{c}{\textbf{Adult}} \\
 \cmidrule(lr){2-4} \cmidrule(lr){5-7} 
 & \makecell[c]{N in \\ bottom 10} &\makecell[c]{N in \\ bottom 20}& \makecell[c]{Average \\ Rank} &\makecell[c]{N in \\ bottom 10}& \makecell[c]{N in \\ bottom 20} & \makecell[c]{Average \\ Rank} \\
\midrule
DemoShapley & \textbf{3} & 7 & 35.4 & \textbf{4} & \textbf{6} & 32.4 \\
Beta-DemoShapley & \textbf{3} & 6& \textbf{24.1} & 2 & 5 & 38.0 \\
CondAcc & 1 & 5 & 47.2 & 3 & \textbf{6} & \textbf{30.2} \\
Influence & 1 & 5 & 45.3 & 3 & \textbf{6} & 31.5 \\
LOO & 2 & 4 & 38.7 & 1 & 4 & 39.9 \\
\bottomrule
\end{tabular}
\end{adjustbox}
\label{flippedlabeltable}
\end{table}
\subsection{Evaluation for Algorithmic Fairness}

\newadd{
Fairness in machine learning refers to reducing systematic differences in model predictions across sensitive groups, such as gender or race. In the context of ICL, fairness can be affected by the demonstrations chosen for the prompt, as biased samples can propagate or even amplify disparities in predictions~\citep{wang2023decodingtrust}. To examine this effect, we designed experiments on the Adult dataset, focusing on income prediction with gender as the sensitive attribute.  

\noindent\textbf{Fairness Metrics.}  
To quantify fairness, we use base rate parity ($b_P$), demographic parity difference ($M_{dpd}$), and equalized odds difference ($M_{eod}$), defined as follows:
To assess the demographic balance (fairness) of the data distribution, we employ the base rate parity (bP) for distribution P. In this equation, $Y$ is the label for prediction and $A$ is the sensitive attribute in the data.
\begin{equation}
\begin{split}
b_P = \mathbb{P}_{(X, Y, A) \sim P_{X Y}}[Y=1 \mid A=1] \\
      - \mathbb{P}_{(X, Y) \sim P_{X Y A}}[Y=1 \mid A=0]
\end{split}
\end{equation}
The \textbf{demographic parity difference} measures the probability of positive prediction conditioned on the sensitive attribute $A$ being 1 and being 0. The equation is shown below.
\begin{equation}
\begin{split}
M_{dpd}=&\left| \mathbb{P}_{(X, Y, A) \sim P_{X Y}}[f(X)=1 \mid A=1] \right.\\
&\left. - \mathbb{P}_{(X, Y, A) \sim P_{X Y}}[f(X)=1 \mid A=0] \right|
\end{split}
\end{equation}
One drawback of $M_{dpd}$ is that it does not consider the ground-truth labels. \textbf{Equalized odds difference} also considers the ground truth label when calculating the demographic gap. The equation is shown below. 
\begin{equation}
    M_{\mathrm{eod}}=\max \left\{M_{T P}, M_{F P}\right\}
\end{equation}

The $M_{T P}$ denotes the true positive equalized odds difference while the $M_{F P}$ denotes the false positive equalized odds difference:

\begin{equation}
    \begin{split}    
    M_{T P}=&\left| \mathbb{P}_{(X, Y, A) \sim P_{X Y}}[f(X)=1 \mid Y=1, A=0] \right.\\
    &\left. -\mathbb{P}_{(X, Y, A) \sim P_{X Y}}[f(X)=1 \mid Y=1, A=1] \right|
    \end{split}
\end{equation}

\begin{equation}
    \begin{split}
    M_{F P}= \left| \mathbb{P}_{(X, Y, A) \sim P_{X Y}}[f(X)=1 \mid Y=0, A=0] \right. \\
    \left. -\mathbb{P}_{(X, Y, A) \sim P_{X Y}}[f(X)=1 \mid Y=0, A=1] \right|
    \end{split}
\end{equation}


\textbf{Experimental Settings.}  
Following \citet{wang2023decodingtrust}, we construct a test dataset with a fixed base rate parity (BPT) of 0.0, ensuring that any observed bias comes from the demonstration set. We vary the base rate parity of candidate demonstrations (BPC) among 0, 0.5, and 1 to study how biased demonstrations influence fairness. Fairness is measured using $M_{dpd}$ and $M_{eod}$, while accuracy is also reported to capture the trade-off between bias and predictive performance. Experiments are conducted in 16-shot and 32-shot settings. 
}

\textbf{Results}
When both the context and test datasets have BPT = 0, DemoShapley and Beta-DemoShapley achieve the highest accuracy with minimal bias. As BPC increases, bias generally rises across all methods. Samples selected by DemoShapley and Beta-DemoShapley consistently exhibit lower bias while maintaining high accuracy compared to other methods.

\subsection{Discussion}
\lu{summarize our findings and highlight the advantages of our approaches}
\newadd{

Our experiments show that our proposed methods provide consistent benefits across a range of settings in ICL. In the prediction efficacy study, high-valued demonstrations were strongly associated with performance gains, while low-valued ones often degraded results. In the OOD generalization experiments, selecting demonstrations with higher values improved transfer across distributions, with Beta-DemoShapley proving especially effective in low-shot settings. For robustness to label noise, mislabeled data were frequently ranked among the lowest-valued examples, making them identifiable without supervision. In the fairness evaluation, demonstrations selected by our methods reduced disparities across demographic groups while maintaining accuracy.  

These results point to several advantages of our approach. By capturing order sensitivity and weighting small-cardinality subsets more heavily, our framework reflects the unique characteristics of ICL. Because it operates entirely at inference time, it remains applicable to closed-source models where fine-tuning is not feasible. Moreover, its grounding in Shapley theory provides a principled basis for fairness, ensuring that each demonstration is evaluated consistently and objectively. Together, these properties make the DemoShapley family methods practical and reliable tools for demonstration evaluation in real-world applications.
}
\begin{table*}[ht]
\small
\caption{Accuracy (ACC), Demographic Parity Difference ($M_{dpd}$), and Equalized Odds Difference ($M_{eod}$) on the Adult Dataset Using 16-shot and 32-shot ICL with different Base Rate Parity (bpc). The Base Rate Parity of the test set (bpt) is fixed at 0.0. 
}
\vspace{-1.0em}
\label{tab:results}
\begin{center}
\begin{adjustbox}{width=0.85\textwidth}
\begin{tabular}{cccccccccc}
\toprule
\multirow{2}{*}{16-shots (ChatGPT)} & \multicolumn{3}{c}{bpc=0} & \multicolumn{3}{c}{bpc=0.5} & \multicolumn{3}{c}{bpc=1} \\
\cmidrule(lr){2-4} \cmidrule(lr){5-7} \cmidrule(lr){8-10}
 & Acc $\uparrow$ & $M_{dpd}$ $\downarrow$ & $M_{eod}$ $\downarrow$ & Acc $\uparrow$ & $M_{dpd}$ $\downarrow$ & $M_{eod}$ $\downarrow$ & Acc $\uparrow$ & $M_{dpd}$ $\downarrow$ & $M_{eod}$ $\downarrow$\\
\midrule
DemoShapley & 0.736 & 0.029 & \textbf{0.016} & 0.748 & \textbf{0.112} & \textbf{0.136} & \textbf{0.741} & \textbf{0.146} & 0.164 \\
Beta-DemoShapley & \textbf{0.739} & \textbf{0.027} & 0.025 & \textbf{0.751} & 0.123 & 0.136 & 0.737 & 0.110 & 0.155 \\
CondAcc & 0.723 & 0.059 & 0.056 & 0.747 & 0.114 & 0.148 & 0.740 & 0.168 & 0.196 \\
Influence & 0.728 & 0.068 & 0.036 & 0.742 & 0.120 & 0.148 & 0.728 & 0.172 & 0.192 \\
LOO & 0.724 & 0.136 & 0.183 & 0.741 & 0.158 & 0.188 & 0.722 & 0.156 & \textbf{0.160} \\
\bottomrule
\end{tabular}
\end{adjustbox}
\end{center}
\begin{center}
\begin{adjustbox}{width=0.85\textwidth}
\begin{tabular}{cccccccccc}
\toprule
\multirow{2}{*}{32-shots (ChatGPT)} & \multicolumn{3}{c}{bpc=0} & \multicolumn{3}{c}{bpc=0.5} & \multicolumn{3}{c}{bpc=1} \\
\cmidrule(lr){2-4} \cmidrule(lr){5-7} \cmidrule(lr){8-10}
 & Acc $\uparrow$ & $M_{dpd}$ $\downarrow$ & $M_{eod}$ $\downarrow$ & Acc $\uparrow$ & $M_{dpd}$ $\downarrow$ & $M_{eod}$ $\downarrow$ & Acc $\uparrow$ & $M_{dpd}$ $\downarrow$ & $M_{eod}$ $\downarrow$\\
\midrule
DemoShapley & \textbf{0.728} & \textbf{0.076} & \textbf{0.144} & 0.730 & 0.094 & 0.124 & \textbf{0.724} & \textbf{0.122} &\textbf{0.132} \\
Beta-DemoShapley & 0.719 & 0.083 & 0.165 & 0.727 & \textbf{0.084} & \textbf{0.114} & 0.719 & 0.147 & 0.204 \\
CondAcc & 0.710 & 0.124 & 0.184 & 0.715 & 0.102 & 0.148 & 0.711 & 0.170 & 0.240 \\
Influence & 0.625 & 0.142 & 0.156 & 0.622 & 0.116 & 0.140 & 0.708 & 0.140 & 0.220 \\
LOO & 0.721 & 0.138 & 0.172 & \textbf{0.733} & 0.234 & 0.308 & 0.709 & 0.214 & 0.234 \\
\bottomrule
\end{tabular}
\end{adjustbox}    
\end{center}
\label{fairnessresult}
\end{table*}

\vspace{-0.5em}
\section{Conclusion}
We introduce DemoShapley and its extension, Beta-DemoShapley, as robust algorithms for selecting demonstrations in ICL. Building on Data Shapley and Beta Shapley, these methods estimate each demonstration’s value by measuring its marginal impact across prompts of different sizes, starting from the zero-shot case. This dynamic approach captures how each example influences performance in different contexts, unlike other influence-based methods that rely on a fixed number of demonstrations. Beta-DemoShapley further enhances this framework by incorporating the Beta distribution, allowing users to assign higher weights to smaller cardinalities, which aligns with ICL’s prompt length and computational constraints. Together, these methods provide a comprehensive and efficient solution for demonstration selection, consistently improving performance, making it ideal for downstream AI applications.

\section{Acknowledgments}
This work is supported by the National Science Foundation (NSF) Grant \#2312862, NSF-Simons
SkAI Institute, NSF CAREER \#2440542, NSF \#2533996, National Institutes of Health (NIH) \#R01AG091762, and a Goolge Research Scholar Award, Amazon Research Award, and Cisco gift grant.

\bibliographystyle{IEEEtranN}
\bibliography{custom}

\end{document}